\documentclass{article}

\usepackage{PRIMEarxiv}
\usepackage{wrapfig}
\usepackage{natbib} 
\usepackage{multirow}
\usepackage[utf8]{inputenc} 
\usepackage[T1]{fontenc}    
\usepackage{hyperref}       
\usepackage{url}            
\usepackage{booktabs}       
\usepackage{amsfonts}       
\usepackage{nicefrac}       
\usepackage{microtype}      
\usepackage{lipsum}
\usepackage{fancyhdr}       
\usepackage{graphicx}       
\graphicspath{{media/}}     
\usepackage{xcolor}
\usepackage{amsmath}
\usepackage{enumitem}
\usepackage{amssymb}
\newtheorem{theorem}{Theorem}[section]

\newtheorem{lemma}[theorem]{Lemma}

\title{IMPROVE: Iterative Model Pipeline Refinement and Optimization Leveraging LLM Experts}

\author{
    Eric Xue \\
    University of Toronto \\
    \texttt{e.xue@mail.utoronto.ca} \\
    \And
    Ke Chen \\
    University of Illinois at Urbana-Champaign \\
    \texttt{kec10@illinois.edu} \\
    \And
    Zeyi Huang \\
    University of Wisconsin-Madison \\
    \texttt{zeyihuang@cs.wisc.edu} \\
    \AND
    Yuyang Ji \\
    New York University \\
    \texttt{yj2669@nyu.edu} \\
    \And
    Haohan Wang \\
    University of Illinois at Urbana-Champaign \\
    \texttt{haohanw@illinois.edu} \\
}

\begin{document}
\maketitle

\begin{abstract}
Large language model (LLM) agents have emerged as a promising solution to automate the workflow of machine learning, but most existing methods share a common limitation: they attempt to optimize entire pipelines in a single step before evaluation, making it difficult to attribute improvements to specific changes. This lack of granularity leads to unstable optimization and slower convergence, limiting their effectiveness. To address this, we introduce Iterative Refinement, a novel strategy for LLM-driven ML pipeline design inspired by how human ML experts iteratively refine models, focusing on one component at a time rather than making sweeping changes all at once. By systematically updating individual components based on real training feedback, Iterative Refinement improves overall model performance. We also provide some theoretical edvience of the superior properties of this Iterative Refinement. Further, we implement this strategy in IMPROVE, an end-to-end LLM agent framework for automating and optimizing object classification pipelines. Through extensive evaluations across datasets of varying sizes and domains, we demonstrate that Iterative Refinement enables IMPROVE to consistently achieve better performance over existing zero-shot LLM-based approaches.
\end{abstract}


\section{Introduction}
Designing high-performing machine learning (ML) pipelines is a complex, multi-step process involving data preprocessing, model selection, and training configuration. Despite the availability of tools such as Weights \& Biases and progress in hyperparameter optimization~\citep{bergstra2012hpo,bergstra2011hpo,snoek2012hpo,springenberg2016hpo,falkner2018hpo} and neural architecture search~\citep{elsken2017nas,kandasamy2018bo}, most automated solutions optimize individual components in isolation and still require expert intervention.

Recent advances in large language models (LLMs) have made it feasible to automate ML pipeline construction end-to-end~\citep{grosnit2024kaggleagent, trirat2024automlagent, hong2024datainterpreter, li2024autokaggle, guo2024dsagent}. These LLM agent frameworks generate training code, reason about design choices, and coordinate pipeline assembly. However, most current systems attempt to optimize all pipeline components simultaneously. This ``global update'' strategy suffers from three drawbacks: difficulty attributing improvements to specific components, lack of stability due to entangled changes, and slower convergence as performance saturates.

We address these limitations with \textbf{Iterative Refinement}, a structured strategy that updates one pipeline component at a time. Inspired by how human practitioners work, our approach evaluates the impact of each change in isolation and only accepts modifications that improve performance. We implement this strategy in \textbf{IMPROVE} (\textbf{I}terative \textbf{M}odel \textbf{P}ipeline \textbf{R}efinement and \textbf{O}ptimization le\textbf{V}eraging LLM \textbf{E}xperts), a multi-agent framework that autonomously builds, evaluates, and refines image classification pipelines.

Our contributions are both empirical and theoretical:
\begin{itemize}
    \item We propose \textbf{Iterative Refinement}, a novel agent coordination strategy for LLM-driven ML systems that improves stability, interpretability, and performance attribution.
    \item We develop \textbf{IMPROVE}, a fully autonomous multi-agent system that applies Iterative Refinement to optimize pipelines across data preprocessing, model architecture, and training configuration.
    \item We establish theoretical guarantees showing that Iterative Refinement offers faster convergence and greater cost-efficiency than global updates under realistic assumptions about search space structure and LLM resource constraints.
    \item We validate IMPROVE across standard benchmarks and real-world Kaggle datasets, demonstrating that it achieves human-competitive performance and consistently outperforms global and zero-shot baselines.
\end{itemize}

\section{Related Works}
\label{sec:related_works}

\textbf{Large Language Models and LLM Agents.}
Large language models (LLMs), such as GPT-3.5 and GPT-4~\citep{openai2024gpt4}, Claude 3.5, Mixtral, and Llama 3, are pre-trained on large-scale corpora and aligned via supervised fine-tuning and reinforcement learning from human feedback (RLHF)~\citep{long2022rlhf}, enabling capabilities in general-purpose reasoning, code generation, and retrieval.

Code-oriented LLMs like Codex~\citep{chen2021codex} and Code Llama extend support for multiple programming languages. Recent general-purpose LLMs also exhibit strong coding abilities without code-specific fine-tuning.

LLMs have been used to build autonomous agents for reasoning and acting, such as ReAct~\citep{yao2023react}, Reflexion~\citep{shinn2023reflexion}, SwiftSage~\citep{lin2023swiftsage}, and AutoGPT~\citep{autogpt}. Multi-agent frameworks~\citep{hong2024metagpt, du2024team, park2023simulacra} further enhance capabilities by assigning specialized roles~\citep{tseng2024roleplay}, which improves knowledge retrieval and planning~\citep{sreedhar2024singlemulti}.

\textbf{Automated Machine Learning.}
AutoML aims to automate components of the ML pipeline such as data preparation, model selection, and hyperparameter tuning~\citep{he2021automl}. Many efforts focus on individual tasks: AutoAugment~\citep{cubuk2019autoaugment}, Faster AutoAugment~\citep{hataya2019faster}, DADA~\citep{li2020dada}, and TrivialAugment~\citep{muller2021trivial} automate data augmentation via search over predefined policies.

Neural architecture search (NAS) identifies optimal model structures. Early methods used heuristic search~\citep{elsken2017nas}, while more recent work incorporates Bayesian optimization and optimal transport~\citep{kandasamy2018bo}. Similarly, hyperparameter optimization (HPO) improves training configurations: random search~\citep{bergstra2012hpo} surpasses grid search~\citep{bergstra2011hpo}, and Bayesian strategies~\citep{snoek2012hpo,springenberg2016hpo,falkner2018hpo} further refine this process by leveraging performance history.

\textbf{AutoML with LLMs.}
LLMs offer a flexible search space for AutoML, inspiring a growing body of work. Early methods such as GPT-NAS~\citep{yu2023gptnas} used fine-tuned LLMs for neural architecture search (NAS) via crossover and mutation strategies. Later, GENIUS~\citep{zheng2023genius} prompted GPT-4 directly for NAS configurations. Beyond architectures, CAAFE~\citep{hollmann2024caafe} automated feature engineering for tabular data, while AutoML-GPT~\citep{zhang2023automlgpt} tackled end-to-end vision and NLP tasks. AutoMMLab~\citep{yang2024autommlab} enabled natural language–driven training pipeline generation for computer vision.

Recent efforts expand LLMs into full agent systems. Agent K~\citep{grosnit2024kaggleagent} automates Kaggle workflows with LLM planning and Bayesian optimization. AutoML-Agent~\citep{trirat2024automlagent} and Data Interpreter~\citep{hong2024datainterpreter} use multi-agent coordination for pipeline refinement via staged reasoning and graph execution. Other systems target domain-specific automation: AutoKaggle~\citep{li2024autokaggle} and DS-Agent~\citep{guo2024dsagent} handle tabular, text, and time series data through libraries and case-based reasoning. LAMBDA~\citep{sun2024lambda} enables human-in-the-loop collaboration with agent-driven code generation and execution.

\section{Method}
\label{sec:method}
Most LLM-based AutoML systems attempt to optimize the entire pipeline in one step, but this global strategy leads to entangled updates, unstable convergence, and poor attribution of improvement. In contrast, we advocate for an iterative, component-wise refinement strategy that mirrors how human experts systematically tune ML pipelines.

We begin by formalizing this approach and proving theoretical guarantees for its convergence and efficiency. We then introduce the IMPROVE framework, a multi-agent system that implements this strategy, and describe key engineering mechanisms that enable robust execution.

\subsection{Iterative Refinement and Its Theoretical Guarantee}

We formalize the ML pipeline as a composition of $k$ sequential components:
\begin{align}
    F^{(t)} = S_k^{(t)} \circ S_{k-1}^{(t)} \circ \cdots \circ S_1^{(t)},
    \label{eq:pipeline}
\end{align}
where $S_i^{(t)} \in \mathcal{S}_i$ denotes the configuration of the $i$-th module (e.g., data augmentation, model architecture, training schedule) at iteration $t$. Given a dataset $D$ and loss function $\mathcal{L}$, the optimization objective is to minimize:
\[
\mathcal{L}(F^{(t)}(D)) = \mathcal{L}(S_k^{(t)} \circ \cdots \circ S_1^{(t)}(D)).
\]

While prior LLM-based systems attempt to jointly optimize all modules at once, effectively updating the entire tuple $(S_1^{(t)}, \ldots, S_k^{(t)})$, this global strategy suffers from high-dimensional search complexity and poor attribution of performance changes. In contrast, \textbf{Iterative Refinement} updates only one module $S_j$ at a time:
\[
F^{(t+1)} = S_k^{(t)} \circ \cdots \circ S_j^{(t+1)} \circ \cdots \circ S_1^{(t)},
\]
where $S_j^{(t+1)}$ is chosen to minimize loss while holding all other components fixed.

This strategy ensures that each update is:
\begin{itemize}
    \item \textbf{Localized}: The impact of each change is directly measurable under the current pipeline context.
    \item \textbf{Monotonic}: By accepting only improvements, we enforce $\mathcal{L}(F^{(t+1)}(D)) \leq \mathcal{L}(F^{(t)}(D))$.
    \item \textbf{Efficient}: The effective search space per iteration is restricted to a single module, reducing variance and computational cost.
\end{itemize}

We now analyze the convergence and efficiency properties of Iterative Refinement (IR) under a formal optimization framework. The pipeline is modeled as Eq.~\ref{eq:pipeline}. 
The goal is to minimize the loss $\mathcal{L}(F^{(t)}(D))$ on a dataset $D$.

\textbf{Assumptions.}
Our analysis relies on the following assumptions, which reflect realistic properties of LLM-driven AutoML systems:

\begin{itemize}
    \item \textbf{A1. Modular Decomposability:} The pipeline consists of $k$ sequential, reconfigurable modules. Each module’s change impacts downstream modules but does not retroactively affect upstream ones. This follows from the ordered, feedforward nature of most ML pipelines.
    
    \item \textbf{A2. Per-Iteration Candidate Set:} At each iteration $t$, the system generates candidate configurations $\mathcal{C}_j^{(t)} \subset \mathcal{S}_j$ for each module $j$. For IR, only one module is updated at a time from its candidate set; for global optimization, the full tuple is sampled from $\mathcal{C}_{\text{global}}^{(t)} \subset \mathcal{S}_1 \times \cdots \times \mathcal{S}_k$.
    
    \item \textbf{A3. Improvement Decay (LLM Saturation):} The expected gain from global (full-pipeline) optimization decays geometrically over time:
\[
\Delta_{\text{global}}^{(t)} \le \alpha \cdot \gamma^t, \quad \text{for } 0 < \gamma < 1.
\]
This reflects the empirical observation that LLM-generated full-pipeline proposals degrade in effectiveness over successive rounds of self-editing. As the pipeline improves, the probability that random full-scale edits yield further gains diminishes. Additionally, LLMs tend to repeat or overwrite past decisions without introducing truly novel configurations~\citep{shinn2023reflexion}, especially in long-horizon tasks. This leads to performance saturation and diminishing returns. Similar trends have been reported in recent agent-based systems~\citep{grosnit2024kaggleagent, hong2024datainterpreter}, where later rounds often plateau unless guided by finer-grained updates.

\item \textbf{A4. Local Update Stability:} The improvement from component-wise updates is bounded below:
\[
\Delta_{\text{local}}^{(t)} \ge \beta > 0.
\]
This models the practical reliability of targeted refinements: when a single module is modified and evaluated in context, the feedback is clearer and more attributable, enabling more effective learning. Empirically, iterative single-component changes have been shown to maintain stable improvement over longer optimization trajectories~\citep{lin2023swiftsage, trirat2024automlagent}. Furthermore, because local edits preserve the rest of the pipeline, they reduce the risk of regression and allow the system to exploit the current solution state effectively—especially when downstream agents can build on past context without unexpected interference.
\end{itemize}

\textbf{Main Result.}
We now establish the convergence advantage of Iterative Refinement over global optimization as below. 

\begin{theorem}[Crossover Point for Local Dominance]
\label{thm:crossover}
Let $f(T)$ denote the cumulative performance gap between local and global strategies after $T$ iterations:
\[
f(T) := \beta \cdot T - \frac{\alpha \cdot \gamma(1 - \gamma^T)}{1 - \gamma}.
\]
Then $f(T)$ is strictly increasing, and there exists a crossover point
\[
T^* := \min \{ T \in \mathbb{N} \mid f(T) \ge 0 \}
\]
after which local optimization outperforms global optimization in cumulative gain:
\[
\sum_{t=1}^T \Delta_{\text{local}}^{(t)} \ge \sum_{t=1}^T \Delta_{\text{global}}^{(t)} \quad \text{for all } T \ge T^*.
\]
\end{theorem}

\paragraph{Resource-Constrained Optimization.}
While convergence rate is critical, real-world LLM-based systems must also operate under strict resource budgets, especially when implemented through multi-agent prompting architectures. Each optimization step incurs a nontrivial cost due to prompt construction, inter-agent communication, and accumulated context length.

To compare global and local optimization in this setting, we analyze their token-level cost structure. Specifically, we decompose the total resource consumption into three components:
\begin{itemize}
    \item $c_0$: the fixed token cost for system initialization, including loading LLM agents, setting global task definitions, and embedding shared context;
    \item $c_1$: the per-step cost of updating a single module (e.g., composing prompts and interpreting responses);
    \item $c_2$: the amortized cost of logging update history per module per step (e.g., appending "agent $i$ modified component $j$ with config $x$").
\end{itemize}

Under this formulation, we quantify the cumulative token cost of performing $k$ optimization iterations under two strategies:
\[
g_{\text{global}}(k) = c_0 + p c_1 + p c_2 (k - 1), \qquad
g_{\text{local}}(k) = c_0 + c_1 + c_2 (k - 1),
\]
where $p$ is the number of modules in the pipeline. Intuitively, global updates modify all $p$ components per step and log full-context history, while local updates affect only one module at a time, leading to more efficient amortization over time.

To compare convergence under equal-cost budgets, we express the ratio of effective update counts (i.e., the number of steps that can be executed within the same token budget):
\[
\frac{N_{\text{global}}}{N_{\text{local}}} = \frac{g_{\text{global}}(k)}{g_{\text{local}}(k)} = \frac{c_0 + px}{c_0 + x}, \quad \text{where } x := c_1 + c_2(k - 1).
\]

We now ask: given this cost gap, can local updates still outperform global updates under a fixed resource budget? 

To assess convergence under a fixed resource budget, we model the optimization objective as a smooth and strongly convex function \( f(x) \), where \( x \in \mathbb{R}^d \) denotes the full pipeline configuration. We start with the following assumptions. 
\begin{itemize}
    \item \( f \) is \( \mu \)-strongly convex: for all \( x, y \), we have 
    \(
    f(y) \ge f(x) + \nabla f(x)^\top (y - x) + \frac{\mu}{2} \|y - x\|^2;
    \)
    \item \( f \) is \( L \)-smooth: its gradient is Lipschitz continuous with constant \( L \), i.e.,
    \(
    \|\nabla f(x) - \nabla f(y)\| \le L \|x - y\|;
    \)
    \item For coordinate-wise updates, the gradient is coordinate-wise Lipschitz with constant \( L_{\max} \), i.e., each partial derivative \( \nabla_i f \) satisfies:
    \(
    |\nabla_i f(x + \delta e_i) - \nabla_i f(x)| \le L_{\max} |\delta|.
    \)
\end{itemize}

Under these assumptions, global (full-pipeline) updates using gradient descent satisfy:
\[
f(x^{(T)}) - f(x^*) \le \left(1 - \frac{\mu}{L} \right)^T \cdot \left( f(x^{(0)}) - f(x^*) \right),
\]
while component-wise updates using Gauss-Southwell coordinate descent satisfy:
\[
f(x^{(T)}) - f(x^*) \le \left(1 - \frac{\mu_1}{L_{\max}} \right)^T \cdot \left( f(x^{(0)}) - f(x^*) \right),
\]
where \( \mu_1 \in \left[\frac{\mu}{p}, \mu\right] \) is the effective strong convexity along coordinate directions (see \cite{richtarik2014iteration}).

\begin{lemma}[Cost-Normalized Superiority Condition]
\label{lem:cost}
Under fixed budget, iterative refinement outperforms global updates if:
\[
\frac{c_0 + x}{c_0 + px} < \log_{1 - \mu/L}\left(1 - \frac{\mu_1}{L_{\max}}\right).
\]
\end{lemma}

\paragraph{Interpretation and Practical Implications.}
Theorem~\ref{thm:crossover} formalizes a phenomenon often observed in practice: global updates deliver quick early gains but rapidly stagnate due to entangled edits, saturation, and model instability. In contrast, Iterative Refinement maintains steady progress, making it the dominant strategy beyond a modest number of iterations. The threshold $T^*$ quantifies this crossover and is empirically small (e.g., $T^* \approx 2.5$ with $\alpha = 0.3$, $\beta = 0.1$, $\gamma = 0.5$).

Lemma~\ref{lem:cost} shows that even when both strategies are constrained to the same token budget (a key constraint in LLM deployment), IR achieves better convergence due to lower per-step overhead and better amortization of memory costs. This advantage is magnified in highly modular systems (large $p$) or when execution spans many steps (large $k$), making IR especially well-suited for multi-agent systems like IMPROVE, which will be introduced below.

\subsection{IMPROVE}
To evaluate the Iterative Refinement strategy, we developed \textbf{IMPROVE}, a multi-agent end-to-end LLM framework that autonomously generates and optimizes an object classification pipeline. 

\begin{figure}[ht]
    \centering
    \includegraphics[width=\textwidth]{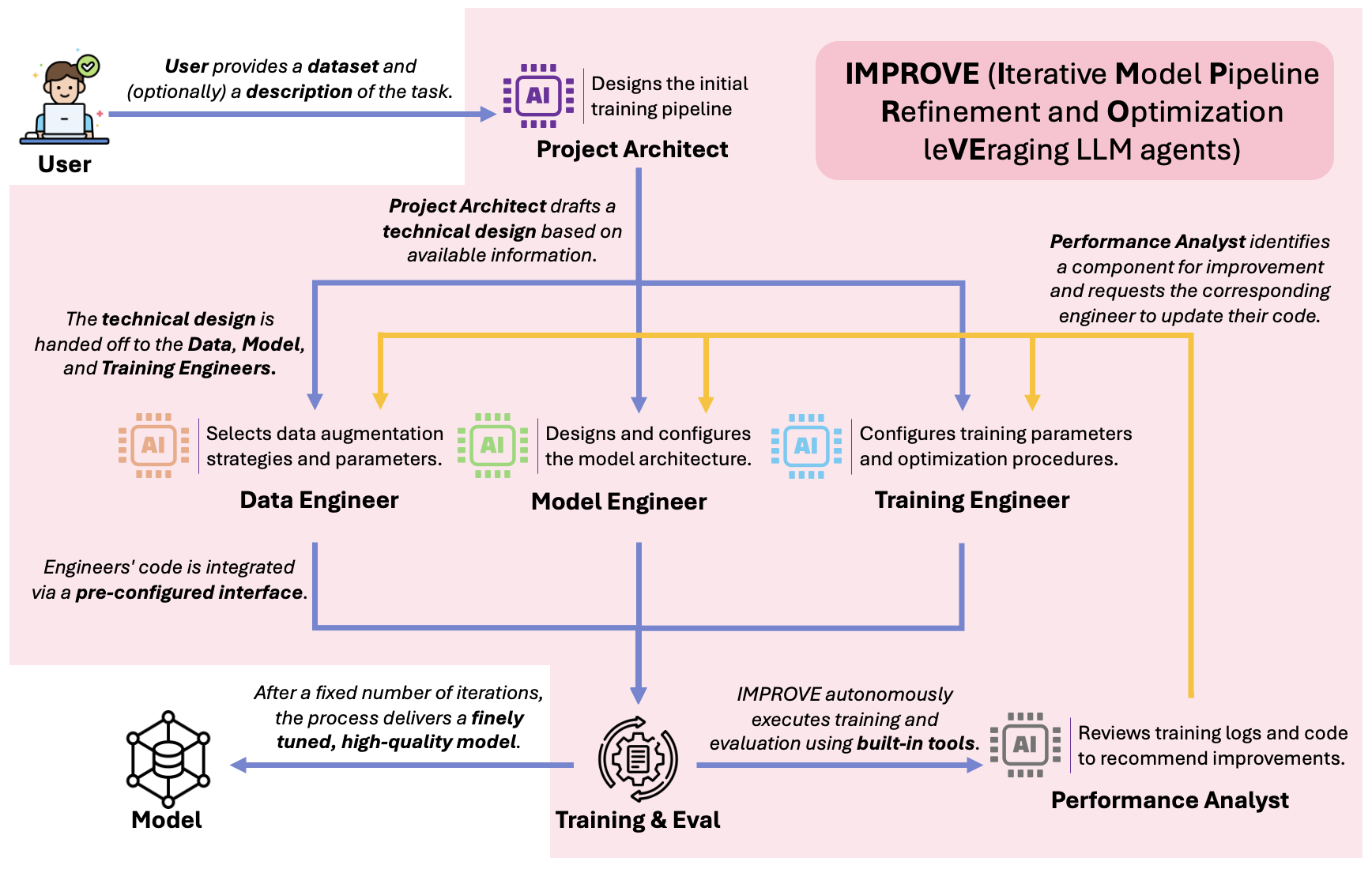}
    \caption{The IMPROVE framework employing Iterative Refinement. Users provide a dataset, and IMPROVE autonomously executes the model development process. The Project Architect designs the pipeline, which is implemented by specialized agents. The Performance Analyst iteratively refines components based on empirical feedback, optimizing the pipeline over multiple iterations. After a set number of iterations, a high-quality model is produced.}
    \label{fig:improve}
\end{figure}

We provide a diagram of the workflow of IMPROVE in Figure~\ref{fig:improve}. It has five specialized roles. 

\begin{itemize}
    \item \textbf{Project Architect}: Analyzes the dataset's size, structure, and metadata to produce a comprehensive technical plan for the full pipeline.
    \item \textbf{Data Engineer}: Implements data processing, including standardization, augmentation, and transformation routines.
    \item \textbf{Model Engineer}: Selects and configures the model architecture, applying modifications and techniques such as regularization as needed.
    \item \textbf{Training Engineer}: Sets hyperparameters (e.g., learning rate, batch size) and chooses appropriate optimization algorithms for model training.
    \item \textbf{Performance Analyst}: Reviews pipeline configurations and training logs, then identifies one area for improvement and issues feedback to a selected engineer.
\end{itemize}

Each agent is aware of its responsibilities, the team’s overall goal, and how to collaborate effectively. Role-specific expertise is embedded through prompt engineering: for instance, the Model Engineer is aware of models like ConvNeXt~\citep{liu2022convnext}, and the Data Engineer leverages advanced augmentations such as MixUp~\citep{zhang2018mixup} and CutMix~\citep{yun2019cutmix}. We explicitly included these techniques in the prompts because, while LLMs are capable of understanding and applying them when instructed, they rarely do so on their own.

IMPROVE applies Iterative Refinement to optimize three primary components of image classification pipelines: \textbf{data augmentation}, \textbf{model architecture}, and \textbf{training procedure}. The workflow begins with the Project Architect generating a technical design from the dataset. This plan guides the Data Engineer, Model Engineer, and Training Engineer in constructing their respective modules. Once the pipeline is assembled, it is trained and evaluated. The Performance Analyst then reviews the results, identifies a single target for refinement, and prompts the corresponding engineer to revise their component. The system retrains and re-evaluates the updated pipeline, retaining only modifications that yield improved performance. Ineffective changes are discarded but logged for traceability. This loop continues for a fixed number of iterations, gradually converging to the best-performing pipeline and model checkpoint.

\paragraph{Engineering Specification}
Beyond Iterative Refinement, IMPROVE integrates several engineering strategies for stable and autonomous pipeline construction.

\textbf{Dataset-Aware Design.}
Users may optionally provide dataset metadata, such as domain, color mode, resolution, or quality, as a natural language description. This information is parsed by the Project Architect and Performance Analyst to inform initial design and diagnostics (see Section~\ref{sec:dataset}).

\textbf{Executable Agent Code.}
Engineer agents generate Python code conforming to predefined I/O interfaces, ensuring composability. IMPROVE parses and integrates these outputs into a unified script, executes the pipeline via shell, and logs results for downstream evaluation.

\textbf{Unified Initialization.}
To improve early-stage stability, the Project Architect generates a complete pipeline in one zero-shot pass before delegating submodules. This yields stronger starting configurations than bottom-up assembly (see Section~\ref{sec:ablation}).

\textbf{Compact Iteration History.}
To avoid context overflow, each refinement step is summarized by an LLM. Only condensed configurations and logs are passed to the Performance Analyst, preserving key signals without exceeding input limits.

\section{Experiments}
\label{sec:experiments}
\subsection{Experimental Setting}
\paragraph{Datasets.} 
We evaluated IMPROVE using two widely recognized classification datasets: CIFAR-10~\citep{krizhevsky2009cifar} and TinyImageNet~\citep{deng2009imagenet}. 

To test IMPROVE's robustness, we also incorporated CIFAR-10-C~\citep{hendrycks2019cifar10c}, a variant of CIFAR-10 designed to assess model performance under common corruptions. CIFAR-10-C introduces 15 corruption types, grouped into four categories: noise, blur, weather, and digital distortions. These corruptions are applied at two severity levels (1 and 5), resulting in 30 distinct versions of the dataset. For experimental efficiency, we created two subsets, CIFAR-10-C-1 and CIFAR-10-C-5, where the suffix denotes the severity level. These subsets were generated by uniformly sampling one image from each corrupted dataset for every test image.

Additionally, we evaluated IMPROVE on two smaller datasets from different domains: SVHN and dSprites. SVHN is a digit classification dataset containing real-world images of house numbers, with variations in resolution and background complexity. dSprites is a synthetic dataset of 2D shapes, with latent factors controlling the shapes’ properties. Specifically, we used the dSprites Orientation dataset, where each shape is rotated into one of 40 possible orientations within the $[0, 2\pi]$ range. Both datasets are part of the Visual Task Adaptation Benchmark (VTAB)~\citep{zhai2020vtab}.

To further assess IMPROVE’s versatility across diverse real-world domains, we also evaluated it on four real-world datasets from Kaggle, a leading platform for machine learning competitions. These datasets span a range of sizes and represent various data domains, including:
\begin{itemize}
    \item \textbf{Cassava Leaf Disease Classification}~\citep{2020cassava}: This dataset consists of 21,367 photos of cassava plants in Uganda, mostly taken using inexpensive cameras. The model must identify whether the plant has one of the four diseases: Mosaic Disease, Green Mottle Brown Streak Disease, Bacterial Blight Disease, or no disease, with 5 classes in total.
    \item \textbf{4 Animal Classification}~\citep{lee2022animal}: This dataset comprises a total of 3,529 images, categorized into four distinct animal classes: cats, deer, dogs, and horses. Each class represents a diverse range of images capturing various poses, environments, and lighting conditions.
    \item \textbf{Arabic Letters Classification}~\citep{khalil2023arabic}: This dataset contains 53,199 images of 65 written Arabic letters, each exhibiting positional variations based on their occurrence within a word with four possible forms: isolated, initial, medial and final, collected from 82 different users.
    \item \textbf{Kitchenware Classification}~\citep{ololo2022kitchenware}: This dataset has 9,367 photos of 6 types of common kitchenware, including cups, glasses, plates, spoons, forks and knives. These photos are taken in households with various lighting and scene conditions.
\end{itemize}

We used three key criteria to guide our selection of Kaggle datasets to ensure an efficient and fair evaluation. First, we excluded datasets with more than a million images to maintain computational feasibility. Second, we only selected datasets used for public competitions, allowing for a direct comparison between IMPROVE's performance and that of human ML practitioners. Third, we chose competitions that used top-1 accuracy as the primary evaluation metric, avoiding those that relied on metrics such as the area under the ROC curve, F1 score, or multiclass log loss, to be consistent across our experiments. After applying these filters, these four datasets were among the few that met all the criteria and aligned with the goals of our evaluation.

We always use the provided test set when available. If no test set is provided, we designate the validation set as the test set. In cases where only a single unsplit dataset is available, we manually split the dataset into training and test sets using an 80-20 ratio.

\textit{Baseline.} 
For VTAB datasets, namely SVHN and dSprites Orientation, we compare our results with that of Visual Prompt Tuning (VPT)~\citep{jia2022vpt}, a widely recognized method in the domain of fine-tuning. For all four Kaggle datasets, we benchmark IMPROVE against the top existing leaderboard submissions. To further assess the practicality of the IMPROVE framework, we compare the accuracy of its generated model pipelines with that of a straightforward zero-shot prompting approach, which represents the most likely method a non-ML expert would use to generate a model. For this comparison, we provide the LLM with a simple prompt in the format: "Generate code for training a model on a dataset with X classes." The generated script is then executed, and the resulting model is trained and evaluated on the test set manually.

\textit{Experimental Setup.}
We used two commercial LLMs, GPT-4o and o1, developed by OpenAI, for all experiments. The experiments were conducted over three trials, with the average accuracy and their standard deviation reported. For all IMPROVE runs, we performed 20 iterations to balance optimization and experimental efficiency. For the VTAB datasets (SVHN and dSprites Orientation), we instructed the Model Engineer to use the Vision Transformer (ViT), specifically the ViT-B/16 model, to ensure a fair comparison with the VPT paper, which utilized the same model as its baseline.

\subsection{Main Results}
The results in Table~\ref{tab:zeroshot} demonstrate that IMPROVE consistently outperforms models generated by zero-shot prompting LLMs using both GPT-4o and o1, confirming its effectiveness as a superior alternative for non-ML experts seeking to train high-performing models. For simpler datasets like CIFAR-10, IMPROVE achieves slightly higher accuracy compared to zero-shot prompting, with the difference being roughly between 3\% to 20\%. However, its strength becomes more evident on more challenging datasets like TinyImageNet and the two variants of CIFAR-10 with corrupted images. On these datasets, IMPROVE at most achieves a roughly 40\% higher accuracy than the zero-shot prompting baseline. 

\begin{table}[ht]
\centering
\caption{Average classification accuracy for IMPROVE-generated models and zeroshot prompting LLMs on four standard datasets. The best accuracy on each dataset is bolded.}
\label{tab:zeroshot}

\begin{tabular}{@{}ccccc@{}}
\toprule
Dataset & IMPROVE (o1) & \multicolumn{1}{l}{Zero-shot (o1)} & IMPROVE (GPT-4o) & Zero-shot (GPT-4o) \\ \midrule
CIFAR-10     & \textbf{0.9825}\textsubscript{±0.0018} & 0.7940\textsubscript{±0.0287} & 0.9626\textsubscript{±0.0311} & 0.9290\textsubscript{±0.0331} \\
CIFAR-10-C-1 & \textbf{0.9621}\textsubscript{±0.0026} & 0.7654\textsubscript{±0.0282} & 0.9476\textsubscript{±0.0158} & 0.5425\textsubscript{±0.1007} \\
CIFAR-10-C-5 & \textbf{0.9557}\textsubscript{±0.0128} & 0.7662\textsubscript{±0.0244} & 0.9422\textsubscript{±0.0038} & 0.6218\textsubscript{±0.1724} \\
TinyImageNet & \textbf{0.8692}\textsubscript{±0.0212} & 0.4630\textsubscript{±0.0520} & 0.7875\textsubscript{±0.0169} & 0.4815\textsubscript{±0.1091} \\ \bottomrule
\end{tabular}
\end{table}

\begin{table}[ht]
\centering
\caption{Average classification accuracy for IMPROVE-generated models and zeroshot prompting LLMs on two VTAB Datasets: SVHN and dSprites Orientation. The table also includes the results from full-parameter fine-tuning (FFT) and Visual Prompt Tuning (VPT) from ~\cite{jia2022vpt} for comparison. The best accuracy on each dataset is bolded.}
\label{tab:vpt}

\begin{tabular}{@{}cccccc@{}}
\toprule
Model                   & Dataset              & IMPROVE                & Zeroshot             & FFT              & VPT                \\ \midrule
\multirow{2}{*}{o1}     & SVHN                 & \textbf{0.9779}\textsubscript{±0.0015} & 0.9513\textsubscript{±0.0027} & 0.8740           & 0.7810               \\
                        & dSprites             & \textbf{0.9667}\textsubscript{±0.0005} & 0.5997\textsubscript{±0.1407} & 0.4670           & 0.4790               \\ \midrule
\multirow{2}{*}{GPT-4o} & SVHN                 & \textbf{0.9695}\textsubscript{±0.0046} & 0.9250\textsubscript{±0.0307} & 0.8740           & 0.7810               \\
                        & dSprites             & \textbf{0.9523}\textsubscript{±0.0074} & 0.6515\textsubscript{±0.2199} & 0.4670           & 0.4790               \\ \bottomrule
\end{tabular}
\end{table}

Another notable distinction between the results of IMPROVE and the zero-shot baseline is the significantly lower standard deviation observed with IMPROVE. While zero-shot results occasionally achieve accuracy closer to that of IMPROVE, they are highly inconsistent, even when using the same prompt. For example, the highest recorded standard deviation was approximately ±17\% for GPT-4o on CIFAR-10-C-5. This inconsistency means that inexperienced users might achieve decent results at times, but they could just as easily get terrible models. In contrast, this variability is lowered with IMPROVE, as all runs eventually converge to a consistently strong performance.

Table~\ref{tab:vpt} further illustrates IMPROVE’s effectiveness by comparing it to Visual Prompt Tuning (VPT)~\citep{jia2022vpt}. IMPROVE not only surpasses VPT but also outperforms the full-parameter fine-tuning baselines reported in the same study. On both datasets, IMPROVE again shows clear improvements over zero-shot prompting, regardless of whether GPT-4o or o1 is used.

\begin{table}[ht]
\centering
\caption{Average classification accuracy and leaderboard metrics for IMPROVE-generated models and zeroshot prompting LLMs on four Kaggle Datasets. The table also includes the leaderboard rank of the IMPROVE result, the highest accuracy among all Kaggle submissions, and the average submission attempts for the top 5 leaderboard positions.}
\label{tab:kaggle}

\begin{tabular}{@{}ccccccc@{}}
\toprule
\multirow{2}{*}{Model} & \multirow{2}{*}{Dataset} & \multirow{2}{*}{IMPROVE} & \multirow{2}{*}{Zero-shot} & \multicolumn{3}{c}{Kaggle Statistics} \\
                       &                          &                           &                             & Rank      & Top Acc. & Top Attempts \\ \midrule
\multirow{4}{*}{o1}     & Cassava Leaf Disease     & 0.8931\textsubscript{±0.0023} & 0.8093\textsubscript{±0.0163} & 1591/3900 & 0.9152   & 98           \\
                        & 4 Animals                & 0.9982\textsubscript{±0.0015} & 0.9506\textsubscript{±0.0323} & 1/221     & 0.9991   & 12           \\
                        & Arabic Letters           & 0.9510\textsubscript{±0.0161} & 0.9162\textsubscript{±0.0051} & 6/177     & 0.9680   & 10           \\
                        & Kitchenware              & 0.9763\textsubscript{±0.0048} & 0.9044\textsubscript{±0.0074} & 31/115    & 0.9958   & 10           \\ \midrule
\multirow{4}{*}{GPT-4o} & Cassava Leaf Disease     & 0.8574\textsubscript{±0.0275} & 0.7748\textsubscript{±0.0552} & 2892/3900 & 0.9152   & 98           \\
                        & 4 Animals                & 0.9518\textsubscript{±0.0330} & 0.9196\textsubscript{±0.0600} & 184/221   & 0.9991   & 12           \\
                        & Arabic Letters           & 0.8403\textsubscript{±0.0824} & 0.5946\textsubscript{±0.3264} & 85/177    & 0.9680   & 10           \\
                        & Kitchenware              & 0.9793\textsubscript{±0.0022} & 0.8581\textsubscript{±0.0956} & 25/115    & 0.9958   & 10           \\ \bottomrule
\end{tabular}
\end{table}

In addition to standard datasets, IMPROVE’s performance on Kaggle competition datasets is presented in Table~\ref{tab:kaggle}. These datasets vary in size, image quality, and domain, offering a more realistic assessment of IMPROVE’s adaptability in real-world scenarios. Across all four datasets, IMPROVE consistently outperforms zero-shot prompting, though its top-1 accuracy remains slightly below the highest-ranking Kaggle leaderboard results. Nevertheless, IMPROVE demonstrates its ability to achieve competitive performance comparable to human ML experts, ranking particularly well on smaller and less complex datasets.

These results were achieved with computational efficiency comparable to that of human experts. Each IMPROVE run was limited to 20 iterations, with more than half of these typically encountering code issues such as undefined variables, wrong package usage, or tensor shape mismatches that cause execution failures. However, these error-containing iterations always terminated quickly, typically within a minute, allowing IMPROVE to proceed to the next iteration with minimal impact on overall efficiency. As a result, fewer than 10 valid iterations are executed and analyzed per run. 

\begin{table}[ht]
\centering
\caption{Average classification accuracy for IMPROVE-generated models, with and without Iterative Refinement (IR), and zeroshot prompting LLMs on CIFAR-10 and TinyImageNet. The best accuracy on each dataset is bolded.}
\label{tab:ir}
\begin{tabular}{@{}ccccc@{}}
\toprule
Model                   & Dataset      & IMPROVE                & (NO IR)                  & Zero-shot             \\ \midrule
\multirow{2}{*}{o1}     & CIFAR-10     & \textbf{0.9825}\textsubscript{±0.0018} & 0.9579\textsubscript{±0.0364} & 0.7940\textsubscript{±0.0287} \\
                        & TinyImageNet & \textbf{0.8692}\textsubscript{±0.0212} & 0.8339\textsubscript{±0.0105} & 0.4630\textsubscript{±0.0520} \\ \midrule
\multirow{2}{*}{GPT-4o} & CIFAR-10     & \textbf{0.9626}\textsubscript{±0.0311} & 0.9610\textsubscript{±0.0223} & 0.9290\textsubscript{±0.0331} \\
                        & TinyImageNet & \textbf{0.7875}\textsubscript{±0.0169} & 0.6202\textsubscript{±0.1908} & 0.4815\textsubscript{±0.1091} \\ \bottomrule
\end{tabular}
\end{table}

This number of valid attempts is comparable to the number of submission attempts made by human practitioners in Kaggle competitions, as shown in Table~\ref{tab:kaggle}. However, it is important to note that human-submitted Kaggle code is typically tested extensively in local environments on a validation dataset before submission to ensure the code contains no errors and can achieve at least a decent performance. As a result, human practitioners likely conduct far more optimization iterations beyond those reflected in the reported submission counts. Despite these constraints, IMPROVE demonstrates performance close to that of top human ML practitioners, which again highlights its potential as a powerful tool for automated model development in real-world conditions.

In Table~\ref{tab:ir}, we evaluate the effectiveness of our core technique: Iterative Refinement. To provide a comparison, we developed an alternate version of IMPROVE in which all pipeline components were allowed to be modified simultaneously by their respective agents in every iteration, rather than focusing on improving one component at a time. 

\begin{figure}[ht]
    \centering
    \includegraphics[width=0.8\textwidth]{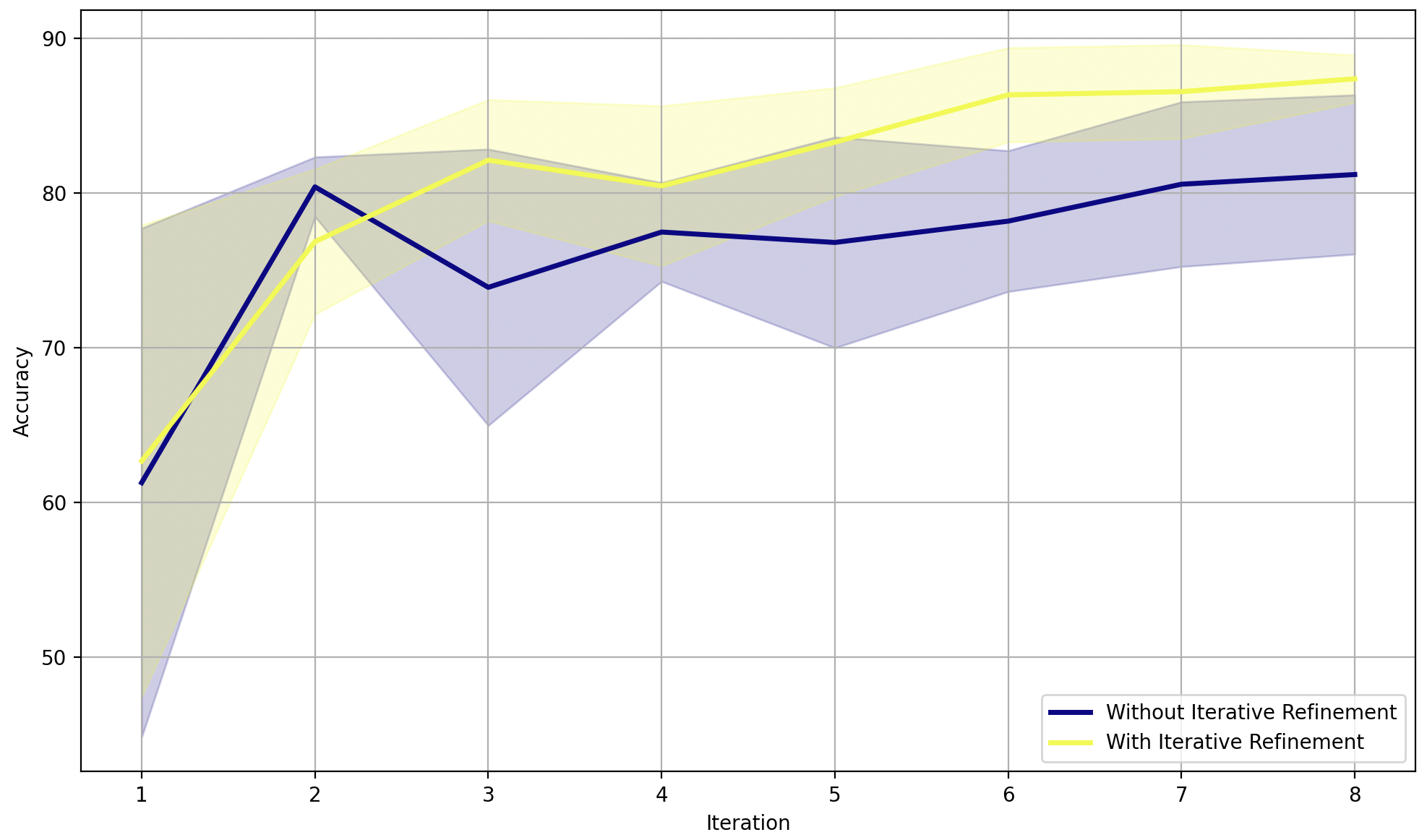} %
    \caption{Average accuracy and standard deviation per iteration for a batch of IMPROVE runs on TinyImageNet using o1. The plot compares the performance trajectory of IMPROVE with Iterative Refinement against a variant where Iterative Refinement is removed. Shaded regions around each line indicate the standard deviation. Iterations that resulted in errors, which do not impact accuracy, are omitted.}
    \label{fig:graph}
\end{figure}

The results indicate that this alternative approach led to less coordinated improvements and significantly higher variability in the outcomes. Interestingly, while some individual runs achieved higher performance than IMPROVE, many others performed worse. Observing the trend in Figure~\ref{fig:graph}, we hypothesize that this variability stems from the increased flexibility of modifying multiple components simultaneously. As shown, the approach not using Iterative Refinement expands the search space, offering greater potential to discover an optimal configuration, resulting in fast improvements in the earlier iterations. However, it also introduces a higher risk of failing to converge on a well-performing configuration within the given time constraints. It can be seen that in the later iterations, the version of IMPROVE without Iterative Refinement struggles to find ways to further improve the system, making blind attempts with similar accuracies, whereas IMPROVE with Iterative Refinement can steadily improve its accuracy over time.

\section{Conclusion}
\label{sec:conclusion}
We introduced Iterative Refinement, a strategy for LLM agent-based pipeline optimization that updates components independently for stable, interpretable improvements. We implemented this in IMPROVE, an end-to-end framework that automates image classification via modular agents. IMPROVE incorporates dataset-aware reasoning, code execution, and unified initialization. Experiments show that it outperforms zero-shot baselines, rivals expert performance, and adapts well to corrupted and diverse datasets, demonstrating its practical effectiveness.

\bibliographystyle{plainnat}
\bibliography{main}

\begin{thebibliography}{48}
\providecommand{\natexlab}[1]{#1}
\providecommand{\url}[1]{\texttt{#1}}
\expandafter\ifx\csname urlstyle\endcsname\relax
  \providecommand{\doi}[1]{doi: #1}\else
  \providecommand{\doi}{doi: \begingroup \urlstyle{rm}\Url}\fi

\bibitem[Bergstra and Bengio(2012)]{bergstra2012hpo}
James Bergstra and Yoshua Bengio.
\newblock Random search for hyper-parameter optimization.
\newblock \emph{J. Mach. Learn. Res.}, 13\penalty0 (null):\penalty0 281–305, feb 2012.
\newblock ISSN 1532-4435.

\bibitem[Bergstra et~al.(2011)Bergstra, Bardenet, Bengio, and K\'{e}gl]{bergstra2011hpo}
James Bergstra, R\'{e}mi Bardenet, Yoshua Bengio, and Bal\'{a}zs K\'{e}gl.
\newblock Algorithms for hyper-parameter optimization.
\newblock In J.~Shawe-Taylor, R.~Zemel, P.~Bartlett, F.~Pereira, and K.Q. Weinberger, editors, \emph{Advances in Neural Information Processing Systems}, volume~24. Curran Associates, Inc., 2011.
\newblock URL \url{https://proceedings.neurips.cc/paper_files/paper/2011/file/86e8f7ab32cfd12577bc2619bc635690-Paper.pdf}.

\bibitem[Chen et~al.(2021)Chen, Tworek, Jun, Yuan, de~Oliveira~Pinto, Kaplan, Edwards, Burda, Joseph, Brockman, Ray, Puri, Krueger, Petrov, Khlaaf, Sastry, Mishkin, Chan, Gray, Ryder, Pavlov, Power, Kaiser, Bavarian, Winter, Tillet, Such, Cummings, Plappert, Chantzis, Barnes, Herbert-Voss, Guss, Nichol, Paino, Tezak, Tang, Babuschkin, Balaji, Jain, Saunders, Hesse, Carr, Leike, Achiam, Misra, Morikawa, Radford, Knight, Brundage, Murati, Mayer, Welinder, McGrew, Amodei, McCandlish, Sutskever, and Zaremba]{chen2021codex}
Mark Chen, Jerry Tworek, Heewoo Jun, Qiming Yuan, Henrique~Ponde de~Oliveira~Pinto, Jared Kaplan, Harri Edwards, Yuri Burda, Nicholas Joseph, Greg Brockman, Alex Ray, Raul Puri, Gretchen Krueger, Michael Petrov, Heidy Khlaaf, Girish Sastry, Pamela Mishkin, Brooke Chan, Scott Gray, Nick Ryder, Mikhail Pavlov, Alethea Power, Lukasz Kaiser, Mohammad Bavarian, Clemens Winter, Philippe Tillet, Felipe~Petroski Such, Dave Cummings, Matthias Plappert, Fotios Chantzis, Elizabeth Barnes, Ariel Herbert-Voss, William~Hebgen Guss, Alex Nichol, Alex Paino, Nikolas Tezak, Jie Tang, Igor Babuschkin, Suchir Balaji, Shantanu Jain, William Saunders, Christopher Hesse, Andrew~N. Carr, Jan Leike, Josh Achiam, Vedant Misra, Evan Morikawa, Alec Radford, Matthew Knight, Miles Brundage, Mira Murati, Katie Mayer, Peter Welinder, Bob McGrew, Dario Amodei, Sam McCandlish, Ilya Sutskever, and Wojciech Zaremba.
\newblock Evaluating large language models trained on code, 2021.
\newblock URL \url{https://arxiv.org/abs/2107.03374}.

\bibitem[Cubuk et~al.(2019)Cubuk, Zoph, Mane, Vasudevan, and Le]{cubuk2019autoaugment}
Ekin~D. Cubuk, Barret Zoph, Dandelion Mane, Vijay Vasudevan, and Quoc~V. Le.
\newblock Autoaugment: Learning augmentation strategies from data.
\newblock In \emph{Proceedings of the IEEE/CVF Conference on Computer Vision and Pattern Recognition (CVPR)}, June 2019.

\bibitem[Deng et~al.(2009)Deng, Dong, Socher, Li, Li, and Fei-Fei]{deng2009imagenet}
Jia Deng, Wei Dong, Richard Socher, Li-Jia Li, Kai Li, and Li~Fei-Fei.
\newblock Imagenet: A large-scale hierarchical image database.
\newblock In \emph{2009 IEEE Conference on Computer Vision and Pattern Recognition}, pages 248--255. IEEE, 2009.

\bibitem[Du et~al.(2024)Du, Qian, Liu, Xie, Wang, Dang, Chen, and Yang]{du2024team}
Zhuoyun Du, Chen Qian, Wei Liu, Zihao Xie, Yifei Wang, Yufan Dang, Weize Chen, and Cheng Yang.
\newblock Multi-agent software development through cross-team collaboration, 2024.
\newblock URL \url{https://arxiv.org/abs/2406.08979}.

\bibitem[Elsken et~al.(2017)Elsken, Metzen, and Hutter]{elsken2017nas}
Thomas Elsken, Jan-Hendrik Metzen, and Frank Hutter.
\newblock Simple and efficient architecture search for convolutional neural networks, 2017.
\newblock URL \url{https://arxiv.org/abs/1711.04528}.

\bibitem[Falkner et~al.(2018)Falkner, Klein, and Hutter]{falkner2018hpo}
Stefan Falkner, Aaron Klein, and Frank Hutter.
\newblock {BOHB}: Robust and efficient hyperparameter optimization at scale.
\newblock In Jennifer Dy and Andreas Krause, editors, \emph{Proceedings of the 35th International Conference on Machine Learning}, volume~80 of \emph{Proceedings of Machine Learning Research}, pages 1437--1446. PMLR, 10--15 Jul 2018.
\newblock URL \url{https://proceedings.mlr.press/v80/falkner18a.html}.

\bibitem[Grosnit et~al.(2024)Grosnit, Maraval, Doran, Paolo, Thomas, Beevi, Gonzalez, Khandelwal, Iacobacci, Benechehab, Cherkaoui, El-Hili, Shao, Hao, Yao, Kegl, Bou-Ammar, and Wang]{grosnit2024kaggleagent}
Antoine Grosnit, Alexandre Maraval, James Doran, Giuseppe Paolo, Albert Thomas, Refinath Shahul Hameed~Nabeezath Beevi, Jonas Gonzalez, Khyati Khandelwal, Ignacio Iacobacci, Abdelhakim Benechehab, Hamza Cherkaoui, Youssef~Attia El-Hili, Kun Shao, Jianye Hao, Jun Yao, Balazs Kegl, Haitham Bou-Ammar, and Jun Wang.
\newblock Large language models orchestrating structured reasoning achieve kaggle grandmaster level, 2024.
\newblock URL \url{https://arxiv.org/abs/2411.03562}.

\bibitem[Guo et~al.(2024)Guo, Deng, Wen, Chen, Chang, and Wang]{guo2024dsagent}
Siyuan Guo, Cheng Deng, Ying Wen, Hechang Chen, Yi~Chang, and Jun Wang.
\newblock Ds-agent: Automated data science by empowering large language models with case-based reasoning, 2024.
\newblock URL \url{https://arxiv.org/abs/2402.17453}.

\bibitem[Hataya et~al.(2019)Hataya, Zdenek, Yoshizoe, and Nakayama]{hataya2019faster}
Ryuichiro Hataya, Jan Zdenek, Kazuki Yoshizoe, and Hideki Nakayama.
\newblock Faster autoaugment: Learning augmentation strategies using backpropagation, 2019.
\newblock URL \url{https://arxiv.org/abs/1911.06987}.

\bibitem[He et~al.(2021)He, Zhao, and Chu]{he2021automl}
Xin He, Kaiyong Zhao, and Xiaowen Chu.
\newblock Automl: A survey of the state-of-the-art.
\newblock \emph{Knowledge-Based Systems}, 212:\penalty0 106622, January 2021.
\newblock ISSN 0950-7051.
\newblock \doi{10.1016/j.knosys.2020.106622}.
\newblock URL \url{http://dx.doi.org/10.1016/j.knosys.2020.106622}.

\bibitem[Hendrycks and Dietterich(2019)]{hendrycks2019cifar10c}
Dan Hendrycks and Thomas Dietterich.
\newblock Benchmarking neural network robustness to common corruptions and perturbations.
\newblock \emph{Proceedings of the International Conference on Learning Representations}, 2019.

\bibitem[Hollmann et~al.(2024)Hollmann, M\"{u}ller, and Hutter]{hollmann2024caafe}
Noah Hollmann, Samuel M\"{u}ller, and Frank Hutter.
\newblock Large language models for automated data science: introducing caafe for context-aware automated feature engineering.
\newblock In \emph{Proceedings of the 37th International Conference on Neural Information Processing Systems}, Red Hook, NY, USA, 2024. Curran Associates Inc.

\bibitem[Hong et~al.(2024{\natexlab{a}})Hong, Lin, Liu, Liu, Wu, Zhang, Wei, Li, Chen, Zhang, Wang, Zhang, Zhang, Yang, Zhuge, Guo, Zhou, Tao, Tang, Lu, Zheng, Liang, Fei, Cheng, Gou, Xu, and Wu]{hong2024datainterpreter}
Sirui Hong, Yizhang Lin, Bang Liu, Bangbang Liu, Binhao Wu, Ceyao Zhang, Chenxing Wei, Danyang Li, Jiaqi Chen, Jiayi Zhang, Jinlin Wang, Li~Zhang, Lingyao Zhang, Min Yang, Mingchen Zhuge, Taicheng Guo, Tuo Zhou, Wei Tao, Xiangru Tang, Xiangtao Lu, Xiawu Zheng, Xinbing Liang, Yaying Fei, Yuheng Cheng, Zhibin Gou, Zongze Xu, and Chenglin Wu.
\newblock Data interpreter: An llm agent for data science, 2024{\natexlab{a}}.
\newblock URL \url{https://arxiv.org/abs/2402.18679}.

\bibitem[Hong et~al.(2024{\natexlab{b}})Hong, Zhuge, Chen, Zheng, Cheng, Wang, Zhang, Wang, Yau, Lin, Zhou, Ran, Xiao, Wu, and Schmidhuber]{hong2024metagpt}
Sirui Hong, Mingchen Zhuge, Jonathan Chen, Xiawu Zheng, Yuheng Cheng, Jinlin Wang, Ceyao Zhang, Zili Wang, Steven Ka~Shing Yau, Zijuan Lin, Liyang Zhou, Chenyu Ran, Lingfeng Xiao, Chenglin Wu, and J{\"u}rgen Schmidhuber.
\newblock Meta{GPT}: Meta programming for a multi-agent collaborative framework.
\newblock In \emph{The Twelfth International Conference on Learning Representations}, 2024{\natexlab{b}}.
\newblock URL \url{https://openreview.net/forum?id=VtmBAGCN7o}.

\bibitem[Jia et~al.(2022)Jia, Tang, Chen, Cardie, Belongie, Hariharan, and Lim]{jia2022vpt}
Menglin Jia, Luming Tang, Bor-Chun Chen, Claire Cardie, Serge Belongie, Bharath Hariharan, and Ser-Nam Lim.
\newblock Visual prompt tuning.
\newblock In \emph{Computer Vision – ECCV 2022: 17th European Conference, Tel Aviv, Israel, October 23–27, 2022, Proceedings, Part XXXIII}, page 709–727, Berlin, Heidelberg, 2022. Springer-Verlag.
\newblock ISBN 978-3-031-19826-7.
\newblock \doi{10.1007/978-3-031-19827-4_41}.
\newblock URL \url{https://doi.org/10.1007/978-3-031-19827-4_41}.

\bibitem[Kandasamy et~al.(2018)Kandasamy, Neiswanger, Schneider, Poczos, and Xing]{kandasamy2018bo}
Kirthevasan Kandasamy, Willie Neiswanger, Jeff Schneider, Barnabas Poczos, and Eric~P Xing.
\newblock Neural architecture search with bayesian optimisation and optimal transport.
\newblock In S.~Bengio, H.~Wallach, H.~Larochelle, K.~Grauman, N.~Cesa-Bianchi, and R.~Garnett, editors, \emph{Advances in Neural Information Processing Systems}, volume~31. Curran Associates, Inc., 2018.
\newblock URL \url{https://proceedings.neurips.cc/paper_files/paper/2018/file/f33ba15effa5c10e873bf3842afb46a6-Paper.pdf}.

\bibitem[Khalil(2023)]{khalil2023arabic}
Youssef Khalil.
\newblock Arabic letters classification, 2023.
\newblock URL \url{https://kaggle.com/competitions/arabic-letters-classification}.

\bibitem[Krizhevsky and Hinton(2009)]{krizhevsky2009cifar}
Alex Krizhevsky and Geoffrey Hinton.
\newblock Learning multiple layers of features from tiny images.
\newblock Technical report, University of Toronto, 2009.

\bibitem[Lee et~al.(2022)Lee, Ahn, Park, and Jin]{lee2022animal}
J~H Lee, JHyun Ahn, Sanguk Park, and Seunghyun Jin.
\newblock 4 animal classification, 2022.
\newblock URL \url{https://kaggle.com/competitions/4-animal-classification}.

\bibitem[Li et~al.(2020)Li, Hu, Wang, Hospedales, Robertson, and Yang]{li2020dada}
Yonggang Li, Guosheng Hu, Yongtao Wang, Timothy Hospedales, Neil~M. Robertson, and Yongxin Yang.
\newblock Dada: Differentiable automatic data augmentation, 2020.
\newblock URL \url{https://arxiv.org/abs/2003.03780}.

\bibitem[Li et~al.(2024)Li, Zang, Ma, Guo, Zheng, Liu, Niu, Wang, Yang, Liu, Zhong, Zhou, Huang, and Zhang]{li2024autokaggle}
Ziming Li, Qianbo Zang, David Ma, Jiawei Guo, Tuney Zheng, Minghao Liu, Xinyao Niu, Yue Wang, Jian Yang, Jiaheng Liu, Wanjun Zhong, Wangchunshu Zhou, Wenhao Huang, and Ge~Zhang.
\newblock Autokaggle: A multi-agent framework for autonomous data science competitions, 2024.
\newblock URL \url{https://arxiv.org/abs/2410.20424}.

\bibitem[Lin et~al.(2023)Lin, Fu, Yang, Brahman, Huang, Bhagavatula, Ammanabrolu, Choi, and Ren]{lin2023swiftsage}
Bill~Yuchen Lin, Yicheng Fu, Karina Yang, Faeze Brahman, Shiyu Huang, Chandra Bhagavatula, Prithviraj Ammanabrolu, Yejin Choi, and Xiang Ren.
\newblock Swiftsage: A generative agent with fast and slow thinking for complex interactive tasks, 2023.
\newblock URL \url{https://arxiv.org/abs/2305.17390}.

\bibitem[Liu et~al.(2022)Liu, Mao, Wu, Feichtenhofer, Darrell, and Xie]{liu2022convnext}
Zhuang Liu, Hanzi Mao, Chao-Yuan Wu, Christoph Feichtenhofer, Trevor Darrell, and Saining Xie.
\newblock A convnet for the 2020s, 2022.
\newblock URL \url{https://arxiv.org/abs/2201.03545}.

\bibitem[Mwebaze et~al.(2020)Mwebaze, Mostipak, Joyce, Elliott, and Dane]{2020cassava}
Ernest Mwebaze, Jesse Mostipak, Joyce, Julia Elliott, and Sohier Dane.
\newblock Cassava leaf disease classification, 2020.
\newblock URL \url{https://kaggle.com/competitions/cassava-leaf-disease-classification}.

\bibitem[Müller and Hutter(2021)]{muller2021trivial}
Samuel~G. Müller and Frank Hutter.
\newblock Trivialaugment: Tuning-free yet state-of-the-art data augmentation, 2021.
\newblock URL \url{https://arxiv.org/abs/2103.10158}.

\bibitem[ololo(2022)]{ololo2022kitchenware}
ololo.
\newblock Kitchenware classification, 2022.
\newblock URL \url{https://kaggle.com/competitions/kitchenware-classification}.

\bibitem[OpenAI et~al.(2024)OpenAI, Achiam, Adler, Agarwal, Ahmad, Akkaya, Aleman, Almeida, Altenschmidt, Altman, Anadkat, Avila, Babuschkin, Balaji, Balcom, Baltescu, Bao, Bavarian, Belgum, Bello, Berdine, Bernadett-Shapiro, Berner, Bogdonoff, Boiko, Boyd, Brakman, Brockman, Brooks, Brundage, Button, Cai, Campbell, Cann, Carey, Carlson, Carmichael, Chan, Chang, Chantzis, Chen, Chen, Chen, Chen, Chen, Chess, Cho, Chu, Chung, Cummings, Currier, Dai, Decareaux, Degry, Deutsch, Deville, Dhar, Dohan, Dowling, Dunning, Ecoffet, Eleti, Eloundou, Farhi, Fedus, Felix, Fishman, Forte, Fulford, Gao, Georges, Gibson, Goel, Gogineni, Goh, Gontijo-Lopes, Gordon, Grafstein, Gray, Greene, Gross, Gu, Guo, Hallacy, Han, Harris, He, Heaton, Heidecke, Hesse, Hickey, Hickey, Hoeschele, Houghton, Hsu, Hu, Hu, Huizinga, Jain, Jain, Jang, Jiang, Jiang, Jin, Jin, Jomoto, Jonn, Jun, Kaftan, Łukasz Kaiser, Kamali, Kanitscheider, Keskar, Khan, Kilpatrick, Kim, Kim, Kim, Kirchner, Kiros, Knight, Kokotajlo, Łukasz Kondraciuk, Kondrich,
  Konstantinidis, Kosic, Krueger, Kuo, Lampe, Lan, Lee, Leike, Leung, Levy, Li, Lim, Lin, Lin, Litwin, Lopez, Lowe, Lue, Makanju, Malfacini, Manning, Markov, Markovski, Martin, Mayer, Mayne, McGrew, McKinney, McLeavey, McMillan, McNeil, Medina, Mehta, Menick, Metz, Mishchenko, Mishkin, Monaco, Morikawa, Mossing, Mu, Murati, Murk, Mély, Nair, Nakano, Nayak, Neelakantan, Ngo, Noh, Ouyang, O'Keefe, Pachocki, Paino, Palermo, Pantuliano, Parascandolo, Parish, Parparita, Passos, Pavlov, Peng, Perelman, de~Avila Belbute~Peres, Petrov, de~Oliveira~Pinto, Michael, Pokorny, Pokrass, Pong, Powell, Power, Power, Proehl, Puri, Radford, Rae, Ramesh, Raymond, Real, Rimbach, Ross, Rotsted, Roussez, Ryder, Saltarelli, Sanders, Santurkar, Sastry, Schmidt, Schnurr, Schulman, Selsam, Sheppard, Sherbakov, Shieh, Shoker, Shyam, Sidor, Sigler, Simens, Sitkin, Slama, Sohl, Sokolowsky, Song, Staudacher, Such, Summers, Sutskever, Tang, Tezak, Thompson, Tillet, Tootoonchian, Tseng, Tuggle, Turley, Tworek, Uribe, Vallone, Vijayvergiya,
  Voss, Wainwright, Wang, Wang, Wang, Ward, Wei, Weinmann, Welihinda, Welinder, Weng, Weng, Wiethoff, Willner, Winter, Wolrich, Wong, Workman, Wu, Wu, Wu, Xiao, Xu, Yoo, Yu, Yuan, Zaremba, Zellers, Zhang, Zhang, Zhao, Zheng, Zhuang, Zhuk, and Zoph]{openai2024gpt4}
OpenAI, Josh Achiam, Steven Adler, Sandhini Agarwal, Lama Ahmad, Ilge Akkaya, Florencia~Leoni Aleman, Diogo Almeida, Janko Altenschmidt, Sam Altman, Shyamal Anadkat, Red Avila, Igor Babuschkin, Suchir Balaji, Valerie Balcom, Paul Baltescu, Haiming Bao, Mohammad Bavarian, Jeff Belgum, Irwan Bello, Jake Berdine, Gabriel Bernadett-Shapiro, Christopher Berner, Lenny Bogdonoff, Oleg Boiko, Madelaine Boyd, Anna-Luisa Brakman, Greg Brockman, Tim Brooks, Miles Brundage, Kevin Button, Trevor Cai, Rosie Campbell, Andrew Cann, Brittany Carey, Chelsea Carlson, Rory Carmichael, Brooke Chan, Che Chang, Fotis Chantzis, Derek Chen, Sully Chen, Ruby Chen, Jason Chen, Mark Chen, Ben Chess, Chester Cho, Casey Chu, Hyung~Won Chung, Dave Cummings, Jeremiah Currier, Yunxing Dai, Cory Decareaux, Thomas Degry, Noah Deutsch, Damien Deville, Arka Dhar, David Dohan, Steve Dowling, Sheila Dunning, Adrien Ecoffet, Atty Eleti, Tyna Eloundou, David Farhi, Liam Fedus, Niko Felix, Simón~Posada Fishman, Juston Forte, Isabella Fulford, Leo
  Gao, Elie Georges, Christian Gibson, Vik Goel, Tarun Gogineni, Gabriel Goh, Rapha Gontijo-Lopes, Jonathan Gordon, Morgan Grafstein, Scott Gray, Ryan Greene, Joshua Gross, Shixiang~Shane Gu, Yufei Guo, Chris Hallacy, Jesse Han, Jeff Harris, Yuchen He, Mike Heaton, Johannes Heidecke, Chris Hesse, Alan Hickey, Wade Hickey, Peter Hoeschele, Brandon Houghton, Kenny Hsu, Shengli Hu, Xin Hu, Joost Huizinga, Shantanu Jain, Shawn Jain, Joanne Jang, Angela Jiang, Roger Jiang, Haozhun Jin, Denny Jin, Shino Jomoto, Billie Jonn, Heewoo Jun, Tomer Kaftan, Łukasz Kaiser, Ali Kamali, Ingmar Kanitscheider, Nitish~Shirish Keskar, Tabarak Khan, Logan Kilpatrick, Jong~Wook Kim, Christina Kim, Yongjik Kim, Jan~Hendrik Kirchner, Jamie Kiros, Matt Knight, Daniel Kokotajlo, Łukasz Kondraciuk, Andrew Kondrich, Aris Konstantinidis, Kyle Kosic, Gretchen Krueger, Vishal Kuo, Michael Lampe, Ikai Lan, Teddy Lee, Jan Leike, Jade Leung, Daniel Levy, Chak~Ming Li, Rachel Lim, Molly Lin, Stephanie Lin, Mateusz Litwin, Theresa Lopez, Ryan
  Lowe, Patricia Lue, Anna Makanju, Kim Malfacini, Sam Manning, Todor Markov, Yaniv Markovski, Bianca Martin, Katie Mayer, Andrew Mayne, Bob McGrew, Scott~Mayer McKinney, Christine McLeavey, Paul McMillan, Jake McNeil, David Medina, Aalok Mehta, Jacob Menick, Luke Metz, Andrey Mishchenko, Pamela Mishkin, Vinnie Monaco, Evan Morikawa, Daniel Mossing, Tong Mu, Mira Murati, Oleg Murk, David Mély, Ashvin Nair, Reiichiro Nakano, Rajeev Nayak, Arvind Neelakantan, Richard Ngo, Hyeonwoo Noh, Long Ouyang, Cullen O'Keefe, Jakub Pachocki, Alex Paino, Joe Palermo, Ashley Pantuliano, Giambattista Parascandolo, Joel Parish, Emy Parparita, Alex Passos, Mikhail Pavlov, Andrew Peng, Adam Perelman, Filipe de~Avila Belbute~Peres, Michael Petrov, Henrique~Ponde de~Oliveira~Pinto, Michael, Pokorny, Michelle Pokrass, Vitchyr~H. Pong, Tolly Powell, Alethea Power, Boris Power, Elizabeth Proehl, Raul Puri, Alec Radford, Jack Rae, Aditya Ramesh, Cameron Raymond, Francis Real, Kendra Rimbach, Carl Ross, Bob Rotsted, Henri Roussez,
  Nick Ryder, Mario Saltarelli, Ted Sanders, Shibani Santurkar, Girish Sastry, Heather Schmidt, David Schnurr, John Schulman, Daniel Selsam, Kyla Sheppard, Toki Sherbakov, Jessica Shieh, Sarah Shoker, Pranav Shyam, Szymon Sidor, Eric Sigler, Maddie Simens, Jordan Sitkin, Katarina Slama, Ian Sohl, Benjamin Sokolowsky, Yang Song, Natalie Staudacher, Felipe~Petroski Such, Natalie Summers, Ilya Sutskever, Jie Tang, Nikolas Tezak, Madeleine~B. Thompson, Phil Tillet, Amin Tootoonchian, Elizabeth Tseng, Preston Tuggle, Nick Turley, Jerry Tworek, Juan Felipe~Cerón Uribe, Andrea Vallone, Arun Vijayvergiya, Chelsea Voss, Carroll Wainwright, Justin~Jay Wang, Alvin Wang, Ben Wang, Jonathan Ward, Jason Wei, CJ~Weinmann, Akila Welihinda, Peter Welinder, Jiayi Weng, Lilian Weng, Matt Wiethoff, Dave Willner, Clemens Winter, Samuel Wolrich, Hannah Wong, Lauren Workman, Sherwin Wu, Jeff Wu, Michael Wu, Kai Xiao, Tao Xu, Sarah Yoo, Kevin Yu, Qiming Yuan, Wojciech Zaremba, Rowan Zellers, Chong Zhang, Marvin Zhang, Shengjia
  Zhao, Tianhao Zheng, Juntang Zhuang, William Zhuk, and Barret Zoph.
\newblock Gpt-4 technical report, 2024.
\newblock URL \url{https://arxiv.org/abs/2303.08774}.

\bibitem[Ouyang et~al.(2022)Ouyang, Wu, Jiang, Almeida, Wainwright, Mishkin, Zhang, Agarwal, Slama, Ray, Schulman, Hilton, Kelton, Miller, Simens, Askell, Welinder, Christiano, Leike, and Lowe]{long2022rlhf}
Long Ouyang, Jeff Wu, Xu~Jiang, Diogo Almeida, Carroll~L. Wainwright, Pamela Mishkin, Chong Zhang, Sandhini Agarwal, Katarina Slama, Alex Ray, John Schulman, Jacob Hilton, Fraser Kelton, Luke Miller, Maddie Simens, Amanda Askell, Peter Welinder, Paul Christiano, Jan Leike, and Ryan Lowe.
\newblock Training language models to follow instructions with human feedback, 2022.
\newblock URL \url{https://arxiv.org/abs/2203.02155}.

\bibitem[Park et~al.(2023)Park, O'Brien, Cai, Morris, Liang, and Bernstein]{park2023simulacra}
Joon~Sung Park, Joseph~C. O'Brien, Carrie~J. Cai, Meredith~Ringel Morris, Percy Liang, and Michael~S. Bernstein.
\newblock Generative agents: Interactive simulacra of human behavior, 2023.
\newblock URL \url{https://arxiv.org/abs/2304.03442}.

\bibitem[Richt{\'a}rik and Tak{\'a}{\v{c}}(2014)]{richtarik2014iteration}
Peter Richt{\'a}rik and Martin Tak{\'a}{\v{c}}.
\newblock Iteration complexity of randomized block-coordinate descent methods for minimizing a composite function.
\newblock \emph{Mathematical Programming}, 144\penalty0 (1):\penalty0 1--38, 2014.

\bibitem[Shinn et~al.(2023)Shinn, Cassano, Berman, Gopinath, Narasimhan, and Yao]{shinn2023reflexion}
Noah Shinn, Federico Cassano, Edward Berman, Ashwin Gopinath, Karthik Narasimhan, and Shunyu Yao.
\newblock Reflexion: Language agents with verbal reinforcement learning, 2023.
\newblock URL \url{https://arxiv.org/abs/2303.11366}.

\bibitem[{Significant Gravitas}(2024)]{autogpt}
{Significant Gravitas}.
\newblock {AutoGPT}, 2024.
\newblock URL \url{https://github.com/Significant-Gravitas/AutoGPT}.
\newblock MIT License.

\bibitem[Snoek et~al.(2012)Snoek, Larochelle, and Adams]{snoek2012hpo}
Jasper Snoek, Hugo Larochelle, and Ryan~P Adams.
\newblock Practical bayesian optimization of machine learning algorithms.
\newblock In F.~Pereira, C.J. Burges, L.~Bottou, and K.Q. Weinberger, editors, \emph{Advances in Neural Information Processing Systems}, volume~25. Curran Associates, Inc., 2012.
\newblock URL \url{https://proceedings.neurips.cc/paper_files/paper/2012/file/05311655a15b75fab86956663e1819cd-Paper.pdf}.

\bibitem[Springenberg et~al.(2016)Springenberg, Klein, Falkner, and Hutter]{springenberg2016hpo}
Jost~Tobias Springenberg, Aaron Klein, Stefan Falkner, and Frank Hutter.
\newblock Bayesian optimization with robust bayesian neural networks.
\newblock In D.~Lee, M.~Sugiyama, U.~Luxburg, I.~Guyon, and R.~Garnett, editors, \emph{Advances in Neural Information Processing Systems}, volume~29. Curran Associates, Inc., 2016.
\newblock URL \url{https://proceedings.neurips.cc/paper_files/paper/2016/file/a96d3afec184766bfeca7a9f989fc7e7-Paper.pdf}.

\bibitem[Sreedhar and Chilton(2024)]{sreedhar2024singlemulti}
Karthik Sreedhar and Lydia Chilton.
\newblock Simulating human strategic behavior: Comparing single and multi-agent llms, 2024.
\newblock URL \url{https://arxiv.org/abs/2402.08189}.

\bibitem[Sun et~al.(2024)Sun, Han, Jiang, Qi, Sun, Yuan, and Huang]{sun2024lambda}
Maojun Sun, Ruijian Han, Binyan Jiang, Houduo Qi, Defeng Sun, Yancheng Yuan, and Jian Huang.
\newblock Lambda: A large model based data agent, 2024.
\newblock URL \url{https://arxiv.org/abs/2407.17535}.

\bibitem[Trirat et~al.(2024)Trirat, Jeong, and Hwang]{trirat2024automlagent}
Patara Trirat, Wonyong Jeong, and Sung~Ju Hwang.
\newblock Automl-agent: A multi-agent llm framework for full-pipeline automl, 2024.
\newblock URL \url{https://arxiv.org/abs/2410.02958}.

\bibitem[Tseng et~al.(2024)Tseng, Huang, Hsiao, Chen, Huang, Meng, and Chen]{tseng2024roleplay}
Yu-Min Tseng, Yu-Chao Huang, Teng-Yun Hsiao, Wei-Lin Chen, Chao-Wei Huang, Yu~Meng, and Yun-Nung Chen.
\newblock Two tales of persona in llms: A survey of role-playing and personalization, 2024.
\newblock URL \url{https://arxiv.org/abs/2406.01171}.

\bibitem[Yang et~al.(2024)Yang, Zeng, Jin, Qian, Luo, and Liu]{yang2024autommlab}
Zekang Yang, Wang Zeng, Sheng Jin, Chen Qian, Ping Luo, and Wentao Liu.
\newblock Autommlab: Automatically generating deployable models from language instructions for computer vision tasks, 2024.
\newblock URL \url{https://arxiv.org/abs/2402.15351}.

\bibitem[Yao et~al.(2023)Yao, Zhao, Yu, Du, Shafran, Narasimhan, and Cao]{yao2023react}
Shunyu Yao, Jeffrey Zhao, Dian Yu, Nan Du, Izhak Shafran, Karthik Narasimhan, and Yuan Cao.
\newblock React: Synergizing reasoning and acting in language models, 2023.
\newblock URL \url{https://arxiv.org/abs/2210.03629}.

\bibitem[Yu et~al.(2023)Yu, Liu, Feng, Tang, and Lv]{yu2023gptnas}
Caiyang Yu, Xianggen Liu, Wentao Feng, Chenwei Tang, and Jiancheng Lv.
\newblock Gpt-nas: Evolutionary neural architecture search with the generative pre-trained model, 2023.
\newblock URL \url{https://arxiv.org/abs/2305.05351}.

\bibitem[Yun et~al.(2019)Yun, Han, Oh, Chun, Choe, and Yoo]{yun2019cutmix}
Sangdoo Yun, Dongyoon Han, Seong~Joon Oh, Sanghyuk Chun, Junsuk Choe, and Youngjoon Yoo.
\newblock Cutmix: Regularization strategy to train strong classifiers with localizable features, 2019.
\newblock URL \url{https://arxiv.org/abs/1905.04899}.

\bibitem[Zhai et~al.(2020)Zhai, Puigcerver, Kolesnikov, Ruyssen, Riquelme, Lucic, Djolonga, Pinto, Neumann, Dosovitskiy, Beyer, Bachem, Tschannen, Michalski, Bousquet, Gelly, and Houlsby]{zhai2020vtab}
Xiaohua Zhai, Joan Puigcerver, Alexander Kolesnikov, Pierre Ruyssen, Carlos Riquelme, Mario Lucic, Josip Djolonga, Andre~Susano Pinto, Maxim Neumann, Alexey Dosovitskiy, Lucas Beyer, Olivier Bachem, Michael Tschannen, Marcin Michalski, Olivier Bousquet, Sylvain Gelly, and Neil Houlsby.
\newblock A large-scale study of representation learning with the visual task adaptation benchmark, 2020.
\newblock URL \url{https://arxiv.org/abs/1910.04867}.

\bibitem[Zhang et~al.(2018)Zhang, Cisse, Dauphin, and Lopez-Paz]{zhang2018mixup}
Hongyi Zhang, Moustapha Cisse, Yann~N. Dauphin, and David Lopez-Paz.
\newblock mixup: Beyond empirical risk minimization, 2018.
\newblock URL \url{https://arxiv.org/abs/1710.09412}.

\bibitem[Zhang et~al.(2023)Zhang, Gong, Wu, Liu, and Zhou]{zhang2023automlgpt}
Shujian Zhang, Chengyue Gong, Lemeng Wu, Xingchao Liu, and Mingyuan Zhou.
\newblock Automl-gpt: Automatic machine learning with gpt, 2023.
\newblock URL \url{https://arxiv.org/abs/2305.02499}.

\bibitem[Zheng et~al.(2023)Zheng, Su, You, Wang, Qian, Xu, and Albanie]{zheng2023genius}
Mingkai Zheng, Xiu Su, Shan You, Fei Wang, Chen Qian, Chang Xu, and Samuel Albanie.
\newblock Can gpt-4 perform neural architecture search?, 2023.
\newblock URL \url{https://arxiv.org/abs/2304.10970}.

\end{thebibliography}

\appendix
\section{Ablation Studies}
\subsection{Ablation Studies}
\label{sec:ablation}
\paragraph{Unified Pipeline Initialization.} 
As described earlier, IMPROVE employs an initialization strategy known as unified pipeline initialization (UPI). In this approach, the Project Architect generates a complete training pipeline in a single step using zero-shot prompting, which is then broken down into individual components. UPI leverages the LLM's ability to generate coherent code and recall previously learned configurations, ensuring that the techniques and parameters used across components are well-aligned. As a result, the generated code is more cohesive, leading to a stronger initialization configuration for the model. In contrast, generating each component sequentially - first creating data augmentation code, passing it to the Model Engineer, and then sharing the data and model code with the Training Engineer - often results in less coherent code and reduced model performance. In Table~\ref{tab:init}, we can see that even though IMPROVE without UPI can still outperform the baseline LLM-generated model, the final accuracy is notably lower than that achieved using UPI. This highlights the importance of a strong initial configuration, which enables IMPROVE to converge to a high-performing model more efficiently.

\begin{table}[ht]
\centering
\caption{Average classification accuracy for IMPROVE-generated models,  with and without unified pipeline initialization (UPI), and zeroshot prompting LLMs on CIFAR-10 and TinyImageNet. The best accuracy on each dataset is bolded.}
\label{tab:init}
\begin{tabular}{@{}ccccc@{}}
\toprule
Model                   & Dataset      & IMPROVE                & (NO UPI) & Zero-shot \\ \midrule
\multirow{2}{*}{o1}     & CIFAR-10     & \textbf{0.9825±0.0018} & 0.9528±0.0267    & 0.7940±0.0287 \\
                        & TinyImageNet & \textbf{0.8692±0.0212} & 0.7974±0.0162    & 0.4630±0.0520 \\ \midrule
\multirow{2}{*}{GPT-4o} & CIFAR-10     & \textbf{0.9626±0.0311} & 0.9476±0.0259    & 0.9290±0.0331 \\
                        & TinyImageNet & \textbf{0.7875±0.0169} & 0.6646±0.0981    & 0.4815±0.1091 \\ \bottomrule
\end{tabular}
\end{table}

\textit{Smaller LLMs.}
In our experiments, we primarily used two high-performing LLMs, GPT-4o and o1, to build our LLM agents. To test IMPROVE's robustness with smaller LLMs, we also evaluated its performance using GPT-4o-mini. GPT-4o-mini is presented by OpenAI as the official successor to GPT-3.5, offering improvements in cost, speed, and computational efficiency compared to GPT-3.5 while maintaining a strong performance.

\begin{table}[ht]
\centering
\caption{Average classification accuracy for IMPROVE-generated models and zeroshot prompting LLMs on CIFAR-10 and TinyImageNet using GPT-4o mini, with results from o1 and GPT-4o for comparison.}
\label{tab:mini}
\begin{tabular}{@{}cccc@{}}
\toprule
Model                        & Dataset      & IMPROVE       & Zero-shot \\ \midrule
\multirow{2}{*}{o1}          & CIFAR-10     & 0.9825±0.0018 & 0.7940±0.0287 \\
                             & TinyImageNet & 0.8692±0.0212 & 0.4630±0.0520 \\ \midrule
\multirow{2}{*}{GPT-4o}      & CIFAR-10     & 0.9626±0.0311 & 0.9290±0.0331 \\
                             & TinyImageNet & 0.7875±0.0169 & 0.4815±0.1091 \\ \midrule
\multirow{2}{*}{GPT-4o-mini} & CIFAR-10     & 0.9590±0.0221 & 0.8364±0.0493 \\
                             & TinyImageNet & 0.6847±0.1226 & 0.5781±0.0673 \\ \bottomrule
\end{tabular}
\end{table}

As shown in Table~\ref{tab:mini}, IMPROVE continues to deliver strong results even when using GPT-4o-mini. Despite its smaller size, the model achieves classification accuracies that are significantly higher than those obtained by zero-shot prompting LLMs. This demonstrates IMPROVE's ability to perform well even with limited budget.

\paragraph{Dataset Information Utilization}
\label{sec:dataset}
We also explored how providing additional dataset-specific information influences the design choices and strategies employed by the LLM agents. Specifically, we hypothesize that having access to dataset details would be particularly beneficial for selecting appropriate data augmentation strategies. Our experimental results support this hypothesis, as IMPROVE consistently selected augmentations well-suited to the characteristics of each dataset.

We examined the augmentation strategies IMPROVE employed for each dataset and the reasoning behind them. When training on SVHN, a dataset composed of real-world images of house numbers, IMPROVE selected the ColorJitter augmentation, justifying its choice by stating: "Color jitter (brightness, contrast, saturation, hue) can help in robustifying the model against variations in lighting conditions." This decision is indeed appropriate in this case, as real-world images often contain lighting inconsistencies that can significantly impact model performance.

Conversely, not all augmentations contribute positively to performance. For instance, in the dSprites Orientation dataset, where classification is based on object orientation, applying RandomHorizontalFlip negatively impacts performance by altering class labels and introducing label noise. Although RandomHorizontalFlip was sometimes included in the initial configuration, IMPROVE’s Performance Analyst identified this issue and suggested: "RandomHorizontalFlip may not be useful for orientation classification tasks," and "Orientation-based tasks will not benefit from horizontal flipping; it could even confuse the model."

These findings suggest that IMPROVE is capable of dataset-aware decision-making. Unlike traditional automated approaches that filter through augmentation policies through trial and error, IMPROVE can dynamically adapt its choices based on dataset characteristics and task requirements. By leveraging dataset-specific insights, IMPROVE not only optimizes the augmentation pipeline more effectively but also accelerates model convergence by avoiding ineffective or detrimental transformations.

\section{Proof}
\subsection*{Proof of Theorem~\ref{thm:crossover}}

Recall the cumulative improvement function:
\[
f(T) := \beta T - \frac{\alpha \gamma (1 - \gamma^T)}{1 - \gamma}.
\]
We aim to show:
\[
\exists T^* \in \mathbb{N} \text{ such that } f(T^*) \ge 0, \text{ and } f(T) \text{ is strictly increasing in } T.
\]

\paragraph{Step 1: Monotonicity.}
We compute the discrete derivative:
\[
f(T+1) - f(T) = \beta - \alpha \cdot \gamma^{T+1}.
\]
Since $\gamma \in (0, 1)$, we know $\gamma^{T+1} \to 0$ as $T \to \infty$. Therefore, for sufficiently large $T$, we have:
\[
f(T+1) - f(T) > 0,
\]
implying that $f(T)$ is strictly increasing in $T$.

\paragraph{Step 2: Crossover point existence.}
We analyze the behavior of $f(T)$ as $T \to \infty$:
\[
\lim_{T \to \infty} f(T) = \beta T - \frac{\alpha \gamma}{1 - \gamma}.
\]
Since $\beta T$ grows linearly and the second term is constant, $f(T) \to \infty$. Moreover, $f(0) = - \frac{\alpha \gamma}{1 - \gamma} < 0$, so by monotonicity, there must exist a minimal $T^* \in \mathbb{N}$ such that $f(T^*) \ge 0$.

\hfill $\blacksquare$

\vspace{1em}
\subsection*{Proof of Lemma~\ref{lem:cost}}

We compare convergence rates under a fixed computational budget.

\paragraph{Step 1: Gradient vs. Coordinate Descent Rates.}
Under strong convexity and smoothness assumptions:
\begin{align*}
\text{GD: } & f(x^{(T)}) - f(x^*) \le \left(1 - \frac{\mu}{L} \right)^{T} (f(x^{(0)}) - f(x^*)), \\
\text{IR (GS): } & f(x^{(T)}) - f(x^*) \le \left(1 - \frac{\mu_1}{L_{\max}} \right)^{T} (f(x^{(0)}) - f(x^*)).
\end{align*}
Let $N_{\text{global}}$ and $N_{\text{local}}$ denote the number of steps under each strategy with equivalent cost.

\paragraph{Step 2: Cost Ratio.}
Total token cost over $k$ iterations:
\begin{align*}
g_{\text{global}}(k) &= c_0 + p c_1 + p c_2(k-1) = c_0 + p x, \\
g_{\text{local}}(k) &= c_0 + c_1 + c_2(k-1) = c_0 + x, \quad \text{where } x := c_1 + c_2(k - 1).
\end{align*}
Therefore:
\[
\frac{N_{\text{global}}}{N_{\text{local}}} = \frac{g_{\text{global}}}{g_{\text{local}}} = \frac{c_0 + p x}{c_0 + x}.
\]

\paragraph{Step 3: Convergence Comparison.}
We want:
\[
\left(1 - \frac{\mu_1}{L_{\max}} \right)^{N_{\text{local}}} < \left(1 - \frac{\mu}{L} \right)^{N_{\text{global}}}.
\]
Taking logarithms and rearranging:
\[
\frac{N_{\text{global}}}{N_{\text{local}}} < \frac{\log \left(1 - \frac{\mu_1}{L_{\max}}\right)}{\log \left(1 - \frac{\mu}{L}\right)} = \log_{1 - \mu / L} \left(1 - \frac{\mu_1}{L_{\max}}\right).
\]

Substituting the cost ratio:
\[
\frac{c_0 + x}{c_0 + p x} < \log_{1 - \mu / L} \left(1 - \frac{\mu_1}{L_{\max}}\right).
\]

\hfill $\blacksquare$

\end{document}